\documentclass[10pt,twocolumn,letterpaper]{article}

\usepackage{iccv}
\usepackage{times}
\usepackage{epsfig}
\usepackage{graphicx}
\usepackage{amsmath}
\usepackage{amssymb}
\usepackage{subcaption}

\usepackage[breaklinks=true,bookmarks=false]{hyperref}

\iccvfinalcopy 

\def\iccvPaperID{****} 

\ificcvfinal\fi

\begin{document}

\title{Boosting High Resolution Image Classification with Scaling-up Transformers: \\
Second Place Solution of the CVPPA2023 Deep Nutrient Deficiency Challenge}

\author{Yi Wang$^{1,2}$\\
  $^1${Data Science in Earth Observation, Technical University of Munich (TUM)}\\
  $^2${Remote Sensing Technology Institute, German Aerospace Center (DLR)}\\
}

\maketitle
\ificcvfinal\thispagestyle{empty}\fi

\begin{abstract}
   We present a holistic approach for high resolution image classification that won second place in the ICCV/CVPPA2023 Deep Nutrient Deficiency Challenge\footnote{https://github.com/wangyi111/cvppa2023-DND-challenge.}. The approach consists of a full pipeline of: 1) data distribution analysis to check potential domain shift, 2) backbone selection for a strong baseline model that scales up for high resolution input, 3) transfer learning that utilizes published pretrained models and continuous fine-tuning on small sub-datasets, 4) data augmentation for the diversity of training data and to prevent overfitting, 5) test-time augmentation to improve the prediction's robustness, and 6) "data soups" that conducts cross-fold model prediction average for smoothened final test results.
\end{abstract}

\section{Introduction}
\label{sec:intro}

The CVPPA2023 "Deep Nutrient Deficiency - Dikopshof - Winter Wheat and Winter Rye" challenge targets at image classification to recognize 7 types of nutrient deficiencies. The challenge dataset, DND-Diko-WWWR \cite{yi4549653non}, is a UAV-based RGB dataset that consists of 3,600 overhead images (1024$\times$1024 pixels) of winter wheat harvested in 2020 and winter rye harvested in 2021 (each 1,800 images). 

The challenge is a standard image classification task, yet there are several key points that have to be carefully considered during the preparation phase. First, the dataset size is small, thus overfitting should be expected and avoided. Second, the images have much bigger size than other common datasets while the details are necessary for recognition, thus baseline models have to be selected suitable for high resolution input. Third, the challenge consists of two sub-datasets which have the same class distribution but have different acquisition time and belong to different crop types.

Based on the above thoughts, we developed our solution with a focus on pretrained backbone selection, strong augmentations and inference post-processing. We first analysed the data distribution between train/test, WW2020/WR2021 and different classes. This helps us shape the solution structure. Then, we decided on using Swin Transformer V2 \cite{liu2022swin} as our baseline model, which was designed to fit high resolution images. We initialize the model with ImageNet pretrained weights, and conduct a two-stage continuous fine-tuning by first training WW2020 and WR2021 together and then fine-tuning both subsets separately. We implemented strong data augmentations to improve the model's robustness and generalizability. During inference, we used test-time augmentation (TTA) and 5-fold "data soups" (cross-fold prediction average) to get a smoothened final results.

\begin{table}[]
\centering
\begin{tabular}{ccccc}
\hline
Rank & User          & WW2020        & WR2021        & mAcc          \\ \hline
1    & UQ\_CV         & 94.0          & 93.2          & 93.6          \\
2    & \textbf{botw} & \textbf{94.9} & \textbf{91.9} & \textbf{93.4} \\
3    & Meteor\_elf    & 94.9          & 91.7          & 93.3          \\ \hline
\end{tabular}
\caption{Top3 entries on the final leader board.}
\vspace{-0.5em}
\end{table}

\section{Methodology}

\subsection{Data distribution analysis}
\label{subsec:distribution}
In the very first step, we conduct a preliminary distribution analysis of the challenge dataset. This includes three parts: 1) train/test distribution, 2) class distribution, 3) WW2020/WR2021 distribution. 

\begin{figure*}
    \centering
    \begin{subfigure}{0.33\linewidth}
    \includegraphics[width=\linewidth]{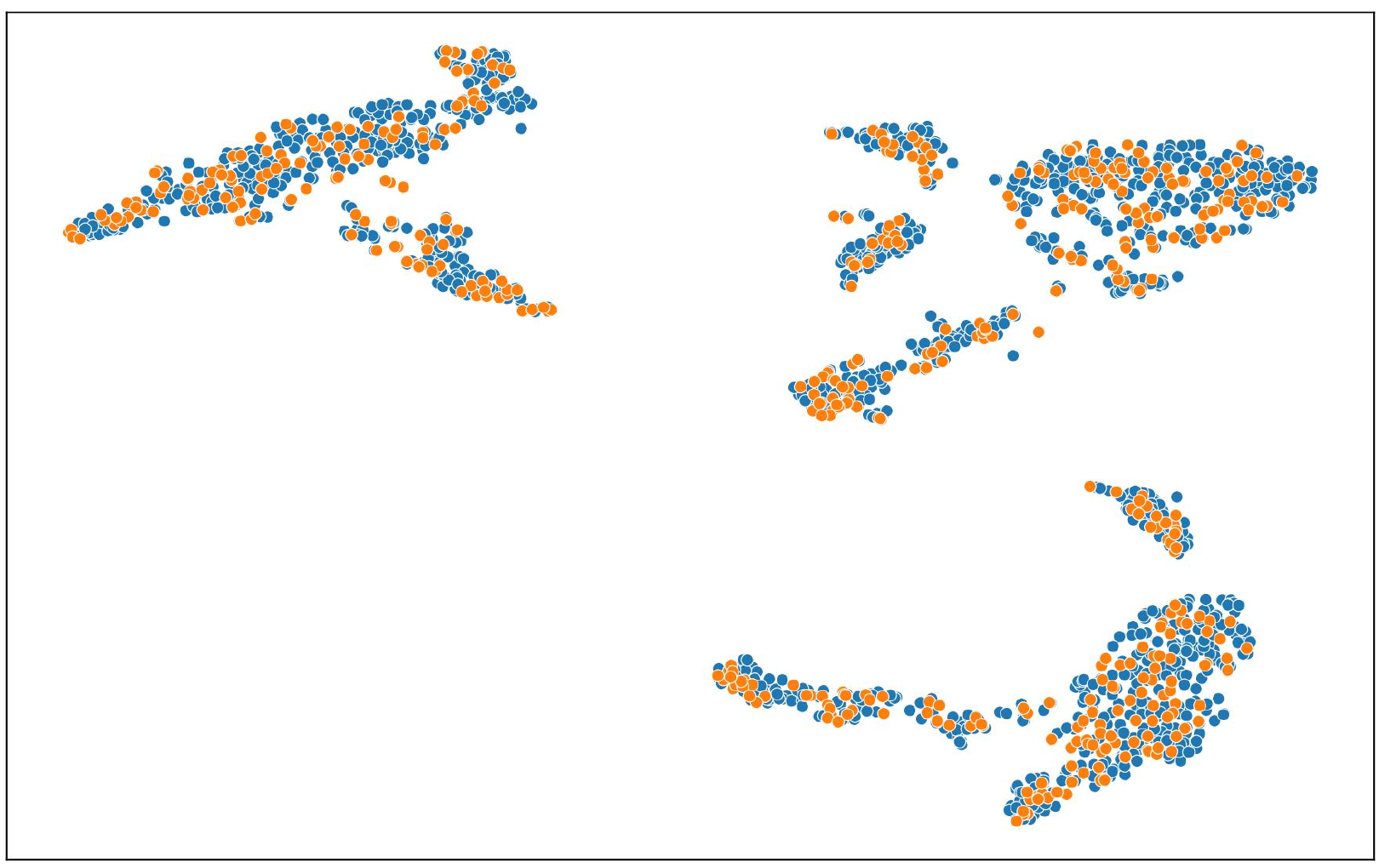}
    \caption{WW2020 train/test distribution.}
    \label{subfig:ww2020}
    \end{subfigure}
    \begin{subfigure}{0.33\linewidth}
    \includegraphics[width=\linewidth]{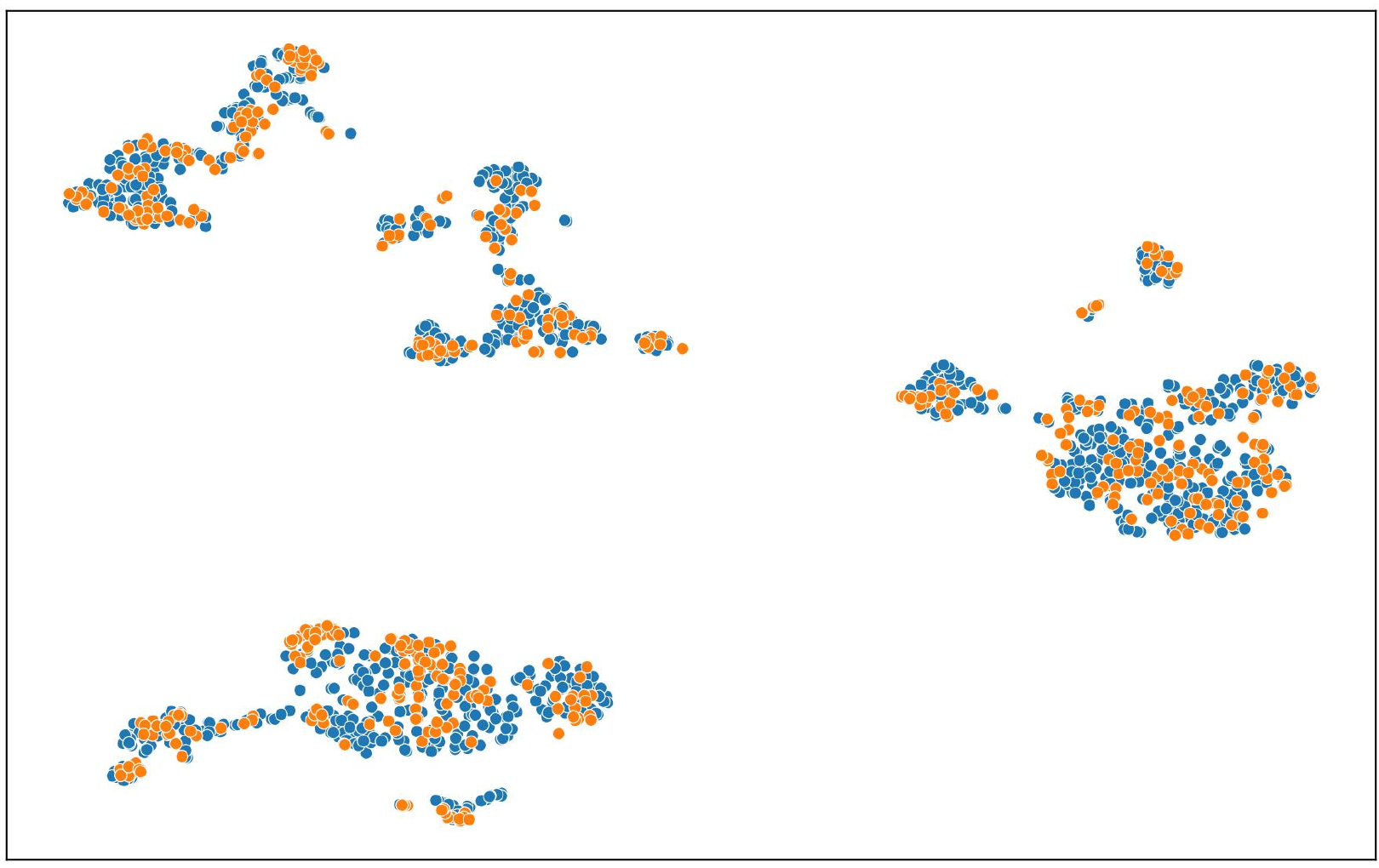}
    \caption{WR2021 train/test distribution.}
    \label{subfig:wr2021}
    \end{subfigure}
    \begin{subfigure}{0.33\linewidth}
    \includegraphics[width=\linewidth]{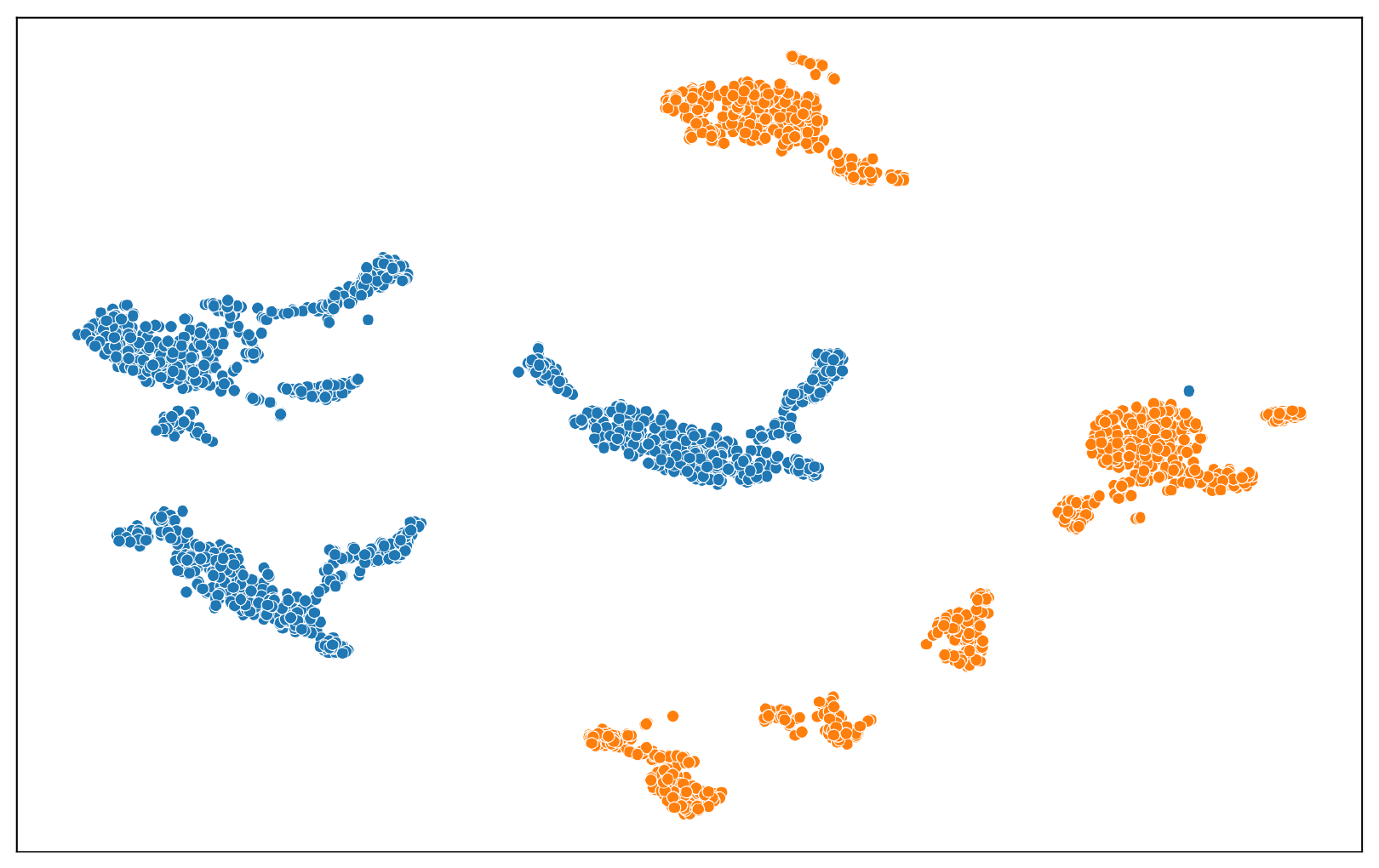}
    \caption{WW2020/WR2021 distribution.}
    \label{subfig:ww2020wr2021}
    \end{subfigure}
    \caption{t-SNE visualization of the challenge datasets. We use ImageNet pretrained ResNet101 to encode 1D features for each image, and then run t-SNE on the features for a rough estimation of the data distribution. The colors blue/orange represent train/test data in $(a)$ and $(b)$, and WW2020/WR2021 data in $(c)$. }
    \label{fig:my_label}
\end{figure*}

The train/test distribution is important to determine if domain shift has to be considered in this challenge. By visual check and a t-SNE \cite{van2008visualizing} visualization of ImageNet-model encoded features, we assume the test set is a random subset of the whole dataset and has a similar distribution as the train set. As can be seen from Figure \ref{subfig:ww2020} and \ref{subfig:wr2021}, the distributions of training and testing data have a strong overlap and can not be easily separated. Therefore, we do not specifically solve domain shift in our solution. 

The class distribution is important to determine if the dataset is balanced among different classes. We calculated the number of images for each class in Table \ref{tab:class-dist}, which shows that the dataset classes are well balanced and we do not need to pay specific attention on the class unbalance.

\begin{table}[h]
\centering
\scalebox{0.63}{
\begin{tabular}{cccccccc}
\hline
       & unfertilized & \_PKCa & N\_KCa & NP\_Ca & NPK\_ & NPKCa & NPKCa+m+s \\ \hline
WW2020 & 192          & 180   & 192   & 192   & 192  & 192   & 192       \\
WR2021 & 192          & 180   & 192   & 192   & 192  & 192   & 192       \\ \hline
\end{tabular}
}
\caption{Number of images for each class.}
\label{tab:class-dist}
\end{table}

The WW2020/WR2021 distribution helps to determine if the two subsets differ from each other and separate training is necessary. By visual check of the two subsets and another t-SNE visualization of ImageNet-model encoded features, we found that they share some similar appearances but the distributions still differ (Figure \ref{subfig:ww2020wr2021}). This indicates that joint training only once would limit the performance, and continuous fine-tuning is a better strategy by training both data together and then fine-tuning separately. This is important since the dataset size is small and training all data together would enlarge the diversity of training data.  In addition, the color/texture distribution of WW2020 looks more complex than WR2021, which may result in performance gap between the two subsets.

\subsection{High resolution model backbone}
A strong baseline model is usually the key for a winning solution in a challenge. Considering the high resolution input and the importance of local details, we decided to use Swin Transformer V2 \cite{liu2022swin} as our model backbone. SwinV2 scales up both capacity and resolution of the original Swin Transformer \cite{liu2021swin}, making it capable of training with very high resolution images (up to 1536$\times$1536).

The official SwinV2-small with window size 8 (we name as SwinV2-s-w8) provided a decent baseline in the early stage of the challenge. We observed that many entries at that stage can not even beat this official baseline, which strengthened our confidence in SwinV2 backbone. As SwinV2-s-w8 was designed for small images, we further extended the model capacity with SwinV2-base-w24 to match the input resolution of 1024$\times$1024.

\subsection{Pretrained model and continuous fine-tuning}
It has been widely proved in practice that pretrained models help improve the efficiency and performance, even when the pretraining domain differs from the target domain. Especially when the model is big but the dataset is small, a good pretrained model is essential for the generalizability. In this challenge, we used the publicly available model weights pretrained on ImageNet\footnote{\url{https://github.com/open-mmlab/mmpretrain/tree/main/configs/swin_transformer_v2}}. 

In addition, though the two sub-datasets in this challenge (WW2020 and WR2021) have different crop types and are in different years, they have a roughly similar distribution as discussed in Section \ref{subsec:distribution}. Therefore, we conduct continuous fine-tuning to make the best use of all the data. Specifically, with the ImageNet pretrained weights, we first train WW2020 and WR2021 together for 50 epochs, and then fine-tune both subsets separately for another 50 epochs.

\begin{figure*}
    \centering
    \includegraphics[width=0.95\linewidth]{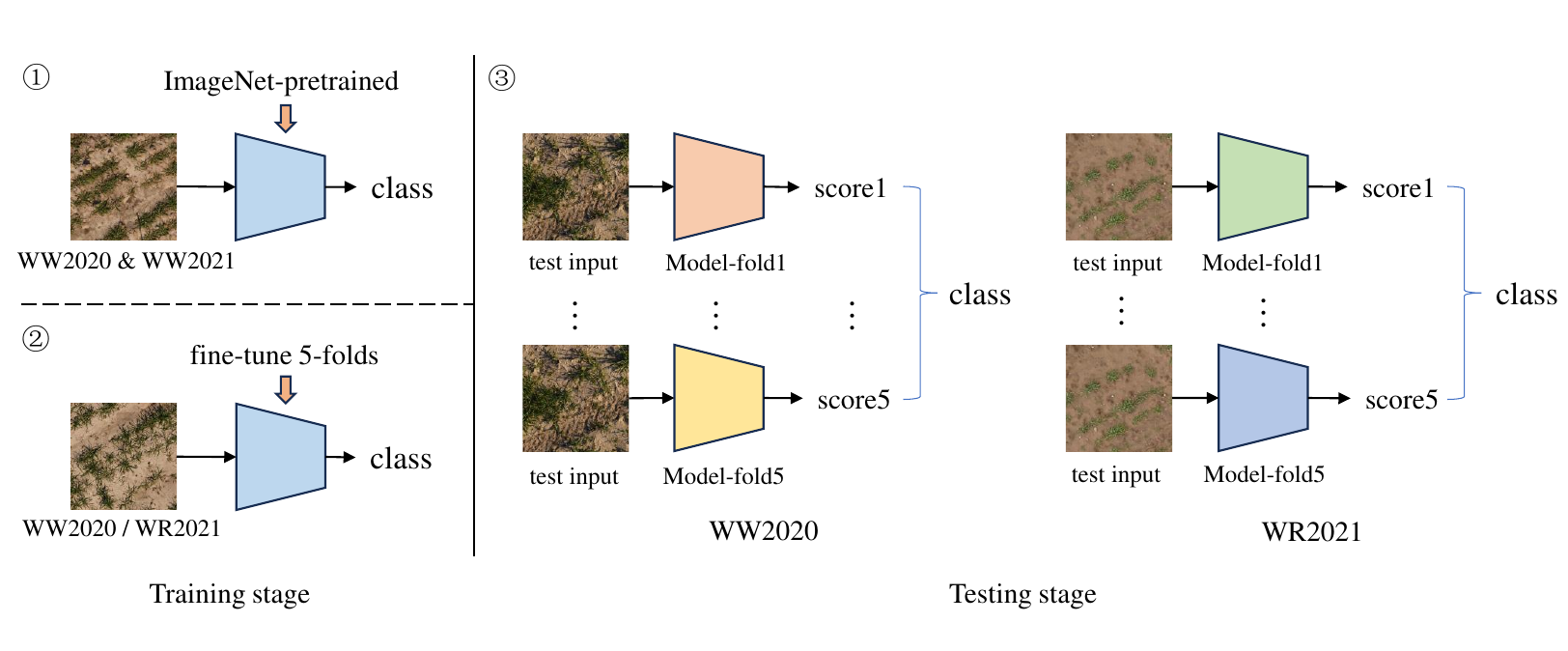}
    \caption{The main workflow of the proposed method. We first transfer ImagNet pre-trained Swin Transformer V2 to train the full challenge data. Then, continuous 5-fold fine-tuning is performed on each subsets separately. In the testing stage, we average model predictions from the 5-fold data soups.}
    \label{fig:workflow}
    \vspace{-1em}
\end{figure*}

\subsection{Data augmentation}
Strong data augmentation plays another big role during training as the dataset size is small and the training data can be easily overfitted. Apart from commonly used "ResizedCrop" and "HorizontalFlip", we added RandAugment \cite{cubuk2020randaugment} and RandomErasing \cite{zhong2020random} to augment the input images. RandAugment is an automated data augmentation method that reduces search space and eliminates a separate proxy task compared to the first automatic augmentation method AutoAugment \cite{cubuk2019autoaugment}. Each time two policies are randomly selected from the policy pool including "AutoContrast", "Equalize", "Invert", "Rotate", "Posterize", "Solarize", "SolarizeAdd", "ColorTransform", "Contrast", "Brightness", "Sharpness", "Shear" and "Translate". RandomErasing works by randomly selecting a small rectangle region and erasing its pixels with random values. 

Furthermore, we added CutMix \cite{yun2019cutmix} and MixUp \cite{zhang2018mixup} as batch augmentations that produce inter-class samples. CutMix randomly cuts out portions of one image and places them over another, and MixUp interpolates the pixel values between two images. Both of these prevent the model from overfitting the training distribution and improve the likelihood that the model can generalize to out of distribution examples.

\subsection{Test-time augmentation}

Test-time augmentation (TTA) is another useful strategy of many challenge winners to improve the performance and reliability of a trained model during the inference or testing phase. TTA involves applying various data augmentation techniques to the input data multiple times and aggregating the model's predictions over these augmented inputs to produce a final prediction. In this challenge, we used "multiscale resize", "resized crop" and "random flip" TTAs.

\subsection{K-fold data soups}
Last but not least, we introduce a "data soups" strategy of averaging K-fold model predictions for post-inference processing. This is implemented by dividing the training data into 5 folds, repeating 5 times training on 4 folds and validating on the remaining fold, inferencing the test data with the 5 models from each fold, and averaging the test predictions (the soup) for final results. 

"data soups" helps prevent the model from overfitting both a specific sub-training set and the full training data, which is a potential issue for traditional "cross-fold validation and then fine-tuning on all data" strategy. The idea is inspired from "model soups" \cite{wortsman2022model}, where the authors proposed to average the weights of models trained with different hyperparameters. In our experiments, we observed "data soups" more beneficial than "model soups".

\section{Implementation details}

\begin{table*}[]
\centering
\scalebox{0.75}{
\begin{tabular}{cccccccc}
\hline
backbone     & pretrain                 & data augmentation                                & test-time augmentation                            & data soups & WW2020        & WR2021        & mAcc          \\ \hline
SwinV2-s-w8  & ImageNet                 & normal (resized crop 896, flip)             & -                              & -          & 79.5          & 84.0          & 81.7          \\
SwinV2-b-w16 & ImageNet                 & normal (resized crop 896, flip)             & -                              & yes        & 88.2          & 90.4          & 89.3          \\
SwinV2-b-w16 & ImageNet + cont. FT & normal (896), CutMix/MixUp                  & flip                           & yes        & 91.9          & 91.5          & 91.7          \\
SwinV2-b-w16 & ImageNet + cont. FT & normal (1024), RandAug, erase, CutMix/MixUp & multiscale, flip               & yes        & 93.4          & 91.7          & 92.5          \\
SwinV2-b-w24 & ImageNet + cont. FT & normal (1024), RandAug, erase, CutMix/MixUp & multiscale, flip, crop & yes        & \textbf{94.9} & \textbf{91.9} & \textbf{93.4} \\ \hline
\end{tabular}
}
\caption{Milestone results during the development. \textit{cont. FT} represents continuous fine-tuning.}
\label{tab:results}
\end{table*}

We used the openmmlab's toolbox mmpretrain \cite{2023mmpretrain} to run the experiments, which is built based on Pytorch. Training was distributed among 4 NVIDIA A100 GPUs and each run took about 1.5 hours. We first trained both WW2020 and WR2021, and then fine tuned each subset separately with the same hyparameters. Implementation details of each subset are described as follows.
\vspace{-1em}
\paragraph{Data} We normalized the pixel values according to ImageNet's mean and standard deviation to better match pretrained weights. For training subset, we used RandomResizedCrop (scale 1024, crop ratio 0.4-1.6), RandomHorizontalFlip, RandAugment ('timm\_increasing' policies, 2 policies at a time, magnitude level 9/10), and RandomErasing (probability 0.25, area ratio 0.01-0.1). For validation subset, we did not use data augmentations. 
\vspace{-1em}
\paragraph{Model} We used SwinV2-base-w24 initialized with ImageNet pretrained weights. The drop path rate was set to 0.2. We conducted CutMix ($\alpha=0.8$) and MixUp ($\alpha=1.0$) for batch-wise augmentions, and a label smoothed cross entropy loss \cite{szegedy2016rethinking}.
\vspace{-1em}
\paragraph{Optimizer} We used AdamW optimizer with learning rate warming up to 1e-5 for 10 epochs, followed by CosineAnnealing decay. We set a total batch size 8 and train each run for 50 epochs. 
\vspace{-1em}
\paragraph{Inference} We divided the training data into 5 folds, trained 5 models, and inferenced the test data with each model. We conducted TTA for each inference, including multiscale resize (1024/1152/1280), random resized crop (crop ratio 0.6-1.4), and random flip. The 5 set of prediction scores were then averaged for final results. 

\section{Results and discussion}

Our solution ranked second in the final leaderboard, with a mean accuracy of 93.4\% averaged between WW2020 (94.9\%) and WR2021 (91.9\%). Table \ref{tab:results} presents some milestone test scores during the development. We observed that the results of WW2020 are generally worse than WR2021 in the early stages while better in the late stages (similar for other entries in the leaderboard). This interesting fact that WW2020 benefits from strong augmentations (in our experiments) potentially reflects our thoughts when analysing the data distribution in Section \ref{subsec:distribution}. In addition, our validation accuracies for WR2021 were almost 100\% for all folds, indicating the possibility of a slight domain shift between training and testing data, which, if solved, might further boost the WR2021 performance.

\section{Conclusion}
In this report, we presented a holistic multi-step approach for high resolution image classification on small datasets. By integrating pre-trained models, continuous fine-tuning, data augmentation, test-time augmentation, and "data soups" prediction average, we boost the test performance upon a strong SwinV2 baseline model. Our solution won the second place in the CVPPA2023 Deep Nutrient Deficiency Challenge, and we hope it could help further studies on computer vision for plant and crop applications.  

\section*{Acknowledgement}
This work is supported by the Helmholtz Association through the Framework of Helmholtz AI. The compute is supported by the Helmholtz Association's Initiative and Networking Fund on the HAICORE@FZJ partition.

{\small
\bibliographystyle{ieee_fullname}
\bibliography{egbib}
}

\end{document}